\documentclass[sigconf,nonacm]{acmart}

\AtBeginDocument{%
  }

\usepackage{booktabs}

\usepackage{amsmath,amssymb}
\usepackage{tikz}
\newcommand{\trip}[3]{\texttt{[#1\allowbreak|#2\allowbreak|#3]}}

\usetikzlibrary{arrows.meta,positioning,fit,calc}
\usepackage{mathtools}
\usepackage{graphicx}
\usepackage{xcolor}
\usepackage{enumitem}
\usepackage{multirow}
\usepackage{array}
\usepackage{url}

\setlist[itemize]{leftmargin=*, topsep=2pt, itemsep=2pt}
\setlist[enumerate]{leftmargin=*, topsep=2pt, itemsep=2pt}

\newcommand{\AUC}{\textsc{auc}}
\newcommand{\TPR}{\textsc{tpr}}
\newcommand{\FPR}{\textsc{fpr}}
\newcommand{\Advantage}{\textsc{adv}}
\newcommand{\TPRatFPR}{\textsc{tpr@fpr}}

\title{Membership Inference Attacks Expose Participation Privacy in ECG Foundation Encoders}

\author{Ziyu Wang}
\authornote{Equal contribution.}
\affiliation{%
  \institution{University of California, Irvine}
  \city{Irvine}
  \state{California}
  \country{United States}
}
\email{ziyuw31@uci.edu}

\author{Elahe Khatibi}
\authornotemark[1]
\affiliation{%
  \institution{University of California, Irvine}
  \city{Irvine}
  \state{California}
  \country{United States}
}
\email{ekhatibi@uci.edu}

\author{Ankita Sharma}
\affiliation{%
  \institution{Arizona State University}
  \city{Tempe}
  \state{Arizona}
  \country{United States}
}
\email{ashar236@asu.edu}

\author{Krishnendu Chakrabarty}
\affiliation{%
  \institution{Arizona State University}
  \city{Tempe}
  \state{Arizona}
  \country{United States}
}
\email{Krishnendu.Chakrabarty@asu.edu}

\author{Sanaz Rahimi Moosavi}
\affiliation{%
  \institution{California State University, Dominguez Hills}
  \city{Carson}
  \state{California}
  \country{United States}
}
\email{srahimimoosavi@csudh.edu}

\author{Farshad Firouzi}
\affiliation{%
  \institution{Arizona State University}
  \city{Tempe}
  \state{Arizona}
  \country{United States}
}
\email{Farshad.Firouzi@asu.edu}

\author{Amir Rahmani}
\affiliation{%
  \institution{University of California, Irvine}
  \city{Irvine}
  \state{California}
  \country{United States}
}
\email{a.rahmani@uci.edu}

\renewcommand{\shortauthors}{Wang and Khatibi et al.}

\keywords{ECG, self-supervised learning, foundation encoders, privacy, membership inference attacks, biosignals}

\begin{document}

\begin{abstract}
Foundation-style ECG encoders pretrained with self-supervised learning are
increasingly reused across tasks, institutions, and deployment contexts,
often through model-as-a-service interfaces that expose scalar scores or latent
representations.
While such reuse improves data efficiency and generalization, it raises a
participation privacy concern: can an adversary infer whether a specific
individual or cohort contributed ECG data to pretraining, even when raw
waveforms and diagnostic labels are never disclosed?
In connected-health settings, training participation itself may reveal
institutional affiliation, study enrollment, or sensitive health context.

We present an implementation-grounded audit of \emph{membership inference
attacks} (MIAs) against modern self-supervised ECG foundation encoders,
covering contrastive objectives (SimCLR, TS2Vec) and masked reconstruction
objectives (CNN- and Transformer-based MAE).
We evaluate three realistic attacker interfaces: (i) score-only black-box
access to scalar outputs, (ii) adaptive learned attackers that aggregate
subject-level statistics across repeated queries, and (iii) embedding-access
attackers that probe latent representation geometry.
Using a subject-centric protocol with window-to-subject aggregation and
calibration at fixed false-positive rates under a cross-dataset auditing
setting, we observe heterogeneous and objective-dependent participation
leakage: leakage is most pronounced in small or institution-specific cohorts
and, for contrastive encoders, can saturate in embedding space, while larger
and more diverse datasets substantially attenuate operational tail risk.
Overall, our results show that restricting access to raw signals or labels is
insufficient to guarantee participation privacy, underscoring the need for
deployment-aware auditing of reusable biosignal foundation encoders in
connected-health systems.
\end{abstract}

\maketitle

\section{Introduction}
Electrocardiograms (ECGs) are among the most widely deployed physiological modalities in connected health~\cite{ebrahimi2020review}.
They are routinely collected in emergency departments and inpatient wards, and increasingly measured through ambulatory and at-home monitoring, supporting applications from rapid clinical triage and arrhythmia screening to long-term risk assessment.
Beyond clinical utility, modern biosignal ecosystems are becoming increasingly \emph{shareable and deployable at scale}: public repositories and community benchmarks such as PhysioNet~\cite{goldberger2000physionet} and PTB-XL~\cite{wagner2020ptbxl} enable reproducible research, while real-world connected-health pipelines rely on cross-institution data exchange across hospitals, vendors, and cloud platforms to improve robustness in heterogeneous populations~\cite{wang2020guardhealth, wang2024differential}.

Motivated by the rapid progress of foundation models in language~\cite{openai2023gpt4,dubey2024llama3,yang2024qwen2}, biosignal modeling is undergoing a similar shift from task-specific supervised training to \emph{foundation-style} representation learning.
In this paradigm, a general-purpose encoder is pretrained on large unlabeled corpora and then reused across downstream tasks, domains, and institutions.
For ECG, recent progress includes contrastive learning of cardiac signals across patients and time (e.g., CLOCS)~\cite{kiyasseh2021clocs}, universal time-series SSL (e.g., TS2Vec)~\cite{yue2022ts2vec}, and masked reconstruction objectives (e.g., MAE)~\cite{he2022mae}.
Importantly, this foundation trend is not unique to ECG: large-scale wearable data have enabled open foundation models for optical physiological signals such as photoplethysmography (PPG), including Pulse-PPG~\cite{pulseppg2025} and PaPaGei~\cite{pillai2025papagei}, alongside industry efforts that train and deploy wearable biosignal foundation models in practice~\cite{apple2024wearablefm,turakhia2019apple}.
Together, these developments make it increasingly common to package pretrained encoders as reusable components accessed through APIs for scoring, monitoring, and downstream personalization.

However, broad reuse of biosignal foundation encoders raises a subtle but consequential privacy question: \emph{does the model reveal whether a specific individual contributed data to pretraining?}
This is a form of participation privacy risk---membership in a training corpus can itself be sensitive, even when raw waveforms and diagnostic labels are never disclosed.
In connected health, membership may imply that a person visited a particular institution, participated in a study, or belongs to a cohort associated with stigmatized conditions, creating downstream harms from discrimination to unwanted health inferences.
Such concerns are amplified for biosignals because physiological traces often contain stable subject-specific morphology and can support biometric recognition~\cite{labati2019deepecg,wang2025linkage, wang2025transecg}.
Crucially, unresolved participation leakage can also degrade trust in clinical AI and discourage institutions and individuals from contributing or sharing data, limiting the scale and diversity of corpora that foundation-style biosignal models rely on for generalization~\cite{gu2021efficiently, wang2024ecg}.

\begin{figure}[t]
    \centering
    \includegraphics[width=\linewidth]{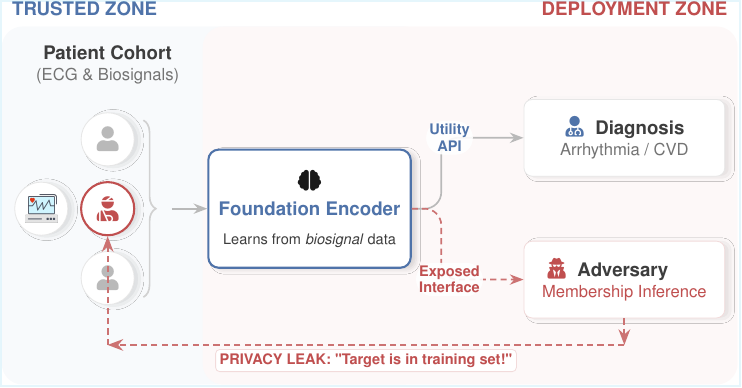}
    \caption{Motivation and threat surface in connected health. A foundation encoder trained on biosignal cohorts may be deployed through exposed interfaces, enabling membership inference and participation privacy leakage.}
    \label{fig:motivation}
\end{figure}

We study this risk through membership inference attacks, which aim to determine whether a target individual was included in a model's training data.
Membership inference has been extensively demonstrated in computer vision and language modeling~\cite{shokri2017membership,yeom2018privacy,carlini2022membership,nasr2019comprehensive}, yet systematic evidence for self-supervised and foundation-style \emph{biosignal} encoders remains limited.
This gap matters because the biosignal setting differs from canonical image/text benchmarks: training data are often cohort-based with relatively few distinct subjects, each contributing many correlated windows, making subject participation potentially more separable.
At the same time, connected-health deployments rarely expose full model internals; they instead expose limited \emph{observable interfaces} such as scalar quality or anomaly scores, or derived statistics aggregated across repeated queries.
Therefore, participation risk in biosignal foundation encoders cannot be assumed from prior domains and must be measured directly under deployment-aligned access constraints.
Figure~\ref{fig:motivation} illustrates this deployment pipeline and its resulting participation-leakage surface.

A key driver of participation leakage is \emph{what the attacker can observe}.
Many realistic deployments return only scalar outputs such as reconstruction errors for quality control, anomaly scores, or similarity-derived metrics, restricting adversaries to score-only inference.
Moreover, practical adversaries can often query multiple windows from the same individual and aggregate evidence across time, which is particularly relevant for ECG where recordings are long and windowed.
Some connected-health systems additionally expose latent representations as
``feature APIs'' to support retrieval, clustering, personalization, or
downstream training.
Such embedding-access interfaces substantially enlarge the attack surface:
representation geometry itself may encode subject-specific stability induced by
pretraining.
In this work, we therefore extend our audit beyond scalar-only interfaces to
explicitly evaluate embedding-space membership inference, modeling deployments
where feature vectors are accessible to external components or partners.
Our evaluation follows a subject-centric protocol with window-to-subject aggregation and calibration at fixed false-positive rates.
Because real-world cohorts are often near-exhaustively used for pretraining, we audit membership using a cross-dataset non-member construction in which non-members are drawn from auxiliary ECG datasets disjoint from the training dataset.
This protocol models a realistic \emph{cross-distribution participation leakage} threat (e.g., across institutions or data sources), and may include dataset-level confounding factors (such as acquisition or resampling artifacts).

We summarize our contributions as follows.
\begin{itemize}
  \item We present an implementation-grounded audit of membership inference against self-supervised ECG foundation encoders across multiple real-world ECG datasets and model families.
  \item We formalize a deployment-grounded participation privacy threat model for connected health, organized by observable interfaces (e.g., scalar scores, aggregated statistics, and
embedding access), and discuss stronger representation-access surfaces as an important future direction.
  \item We propose an operational, subject-centric auditing protocol calibrated at fixed \FPR\ operating points and report standardized leakage metrics (\AUC, \TPRatFPR, and attacker advantage), revealing objective- and cohort-dependent participation leakage under a cross-dataset auditing protocol.
\end{itemize}

\section{Background and Related Work}

\paragraph{Foundation-style representation learning for ECG and biosignals.}
Inspired by the rapid progress of foundation models in language~\cite{openai2023gpt4,dubey2024llama3,yang2024qwen2, wang2025healthq, wang2025medcot}, ECG modeling is shifting from task-specific supervised learning to \emph{foundation-style} representation learning, where general-purpose encoders are pretrained on large unlabeled corpora and reused across tasks and institutions.
This shift is enabled by public resources such as PhysioNet~\cite{goldberger2000physionet} and PTB-XL~\cite{wagner2020ptbxl}, and accelerated by modern SSL objectives.
Contrastive frameworks (e.g., SimCLR~\cite{chen2020simclr}) have inspired ECG-specific methods such as CLOCS~\cite{kiyasseh2021clocs} and general time-series SSL such as TS2Vec~\cite{yue2022ts2vec}, while masked reconstruction (e.g., MAE~\cite{he2022mae}) provides an alternative route to transferable encoders.
Foundation pretraining is also expanding beyond ECG to wearable biosignals, including open PPG foundation models trained across lab and field settings (Pulse-PPG)~\cite{pulseppg2025} and optical physiological foundation models such as PaPaGei~\cite{pillai2025papagei}, alongside large-scale industry efforts trained on wearable biosignals for deployment~\cite{apple2024wearablefm,turakhia2019apple}.
While these models improve data efficiency and generalization, they also concentrate sensitive information into reusable latent spaces, raising privacy risks when encoders are shared widely.

\paragraph{Privacy and identity risks in biosignals and connected health.}
Privacy risks in biosignals extend beyond explicit identifiers~\cite{wang2024ecg, wang2025transecg}.
ECG waveforms capture stable subject-specific morphology and can be leveraged for biometric recognition, including deep-learning-based identification from ECG features~\cite{labati2019deepecg, wang2025linkage}.
In real-world connected health systems, ECG data are often collected longitudinally and exchanged across hospitals, vendors, and cloud platforms to support monitoring, personalization, and large-scale model development~\cite{aqajari2024enhancing, wang2020guardhealth}.
This ecosystem makes identity-centric threats more realistic: even after de-identification, records can be re-identified or linked across datasets through physiological signatures, repeated measurements over time, and cross-release correlations~\cite{wang2025linkage, datta2025react}.
At the same time, the increasing reliance on pretrained representations and reusable encoders shifts the privacy boundary from \emph{raw signal release} to \emph{model-level leakage}---what information can be inferred from a deployed model or its representations.
This perspective aligns with broader concerns in healthcare privacy and data sharing, where participation and linkage risks can carry sensitive implications even when direct identifiers are removed~\cite{ghazarian2021increased}.

\paragraph{Membership inference and participation privacy auditing.}
MIAs aim to infer whether a specific record was used to train a model, providing a concrete auditing lens for \emph{participation privacy}.
Early work introduced shadow-model based MIAs against supervised classifiers~\cite{shokri2017membership}, followed by loss-based attacks that relate membership advantage to generalization gaps~\cite{yeom2018privacy}.
Subsequent work demonstrated that MIAs can remain effective under stronger adversaries and more realistic threat models, including white-box access to model gradients and internals~\cite{nasr2019comprehensive}.
More recently, auditing studies have revisited membership leakage for modern large models, emphasizing that memorization and training dynamics can create measurable privacy risk even without explicit overfitting~\cite{carlini2022membership}.
However, biosignal foundation encoders differ from canonical vision/NLP settings in important ways: datasets are often cohort-based with relatively few distinct subjects, each subject contributes many correlated windows, and deployed interfaces may expose scalar scores or latent embeddings rather than class probabilities.
These differences motivate implementation-grounded evaluation tailored to ECG SSL pretraining.

\paragraph{Positioning of this work.}
Prior biosignal privacy work has largely studied \emph{data-level} threats such as re-identification and linkage on shared records, whereas we focus on an underexplored risk: \emph{model-level participation leakage} from foundation ECG encoders reused across tasks and institutions.
This reuse creates new attack surfaces where an adversary can infer whether a specific individual contributed to pretraining by probing a deployed encoder or its representations, even without raw waveforms.
We provide the first systematic study of membership inference for ECG foundation encoders, formalizing a connected-health threat model. Our results reveal heterogeneous and access-dependent leakage that can degrade trust and discourage data sharing in connected health.

\section{Threat Model and Problem Definition}
\label{sec:threat}

\begin{figure*}[t!]
    \centering
    \includegraphics[width=\linewidth]{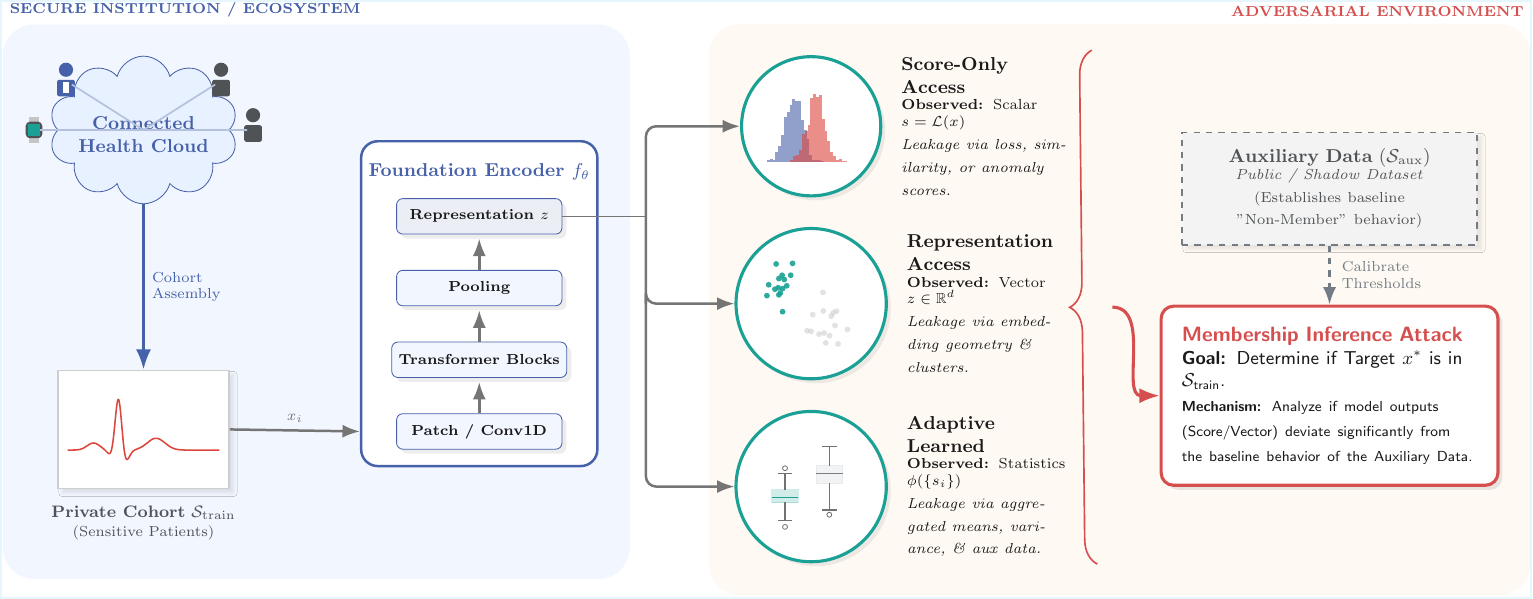}
    \caption{Participation privacy threat model in a connected-health ecosystem.
    A private patient cohort ($\mathcal{S}_{\text{train}}$) is aggregated to train
    a foundation ECG encoder ($f_\theta$), which is subsequently deployed through
    a utility interface.
    An adversary probes the deployed encoder via observable outputs (e.g., scalar
    scores or aggregated statistics) and auxiliary public data
    ($\mathcal{S}_{\text{aux}}$) to infer whether a target subject contributed to
    pretraining.}
    \label{fig:mia_threat}
\end{figure*}

Figure~\ref{fig:mia_threat} illustrates the connected-health threat surface we
study.
A foundation ECG encoder pretrained on private patient cohorts is deployed as a
reusable component for downstream diagnosis, monitoring, or analytics.
An adversary interacts with this deployed encoder through its exposed interface
and attempts to infer whether a target individual or cohort participated in the
pretraining process, resulting in a participation privacy leak.

\paragraph{Pretraining corpus and participation.}
Let a pretraining corpus consist of a set of subjects
$\mathcal{S}=\{s_i\}$, where each subject $s$ contributes a set of ECG windows
$\{x_{s,j}\}_{j=1}^{n_s}$ generated by a fixed and deterministic preprocessing
pipeline.
A target encoder $f_\theta(\cdot)$ is pretrained on a subset
$\mathcal{S}_{\text{train}}\subseteq\mathcal{S}$ and subsequently reused across
tasks and deployment contexts.
Given a query subject $s^\star$, the adversary aims to infer the participation
indicator
\begin{equation*}
y(s^\star)=\mathbb{I}[s^\star \in \mathcal{S}_{\text{train}}].
\end{equation*}
This captures \emph{participation privacy}: even when raw waveforms and clinical
labels are not exposed, membership in a training cohort can itself be sensitive,
as it may reveal institutional affiliation, study participation, or cohort
membership.

For auditing, we obtain ground-truth participation labels from training artifacts
recorded by the pretraining pipeline (e.g., logged subject identifiers).
This enables clean evaluation of membership inference without assuming oracle
access by the attacker.

\paragraph{Observable attacker interface.}
We characterize the adversary by the \emph{observable interface} of the deployed
encoder, reflecting realistic connected-health deployments.
We consider three interface types:

(i) \textbf{Score-only interface}: the deployment exposes only scalar outputs
(e.g., reconstruction error, anomaly score, or quality metric);

(ii) \textbf{Aggregated-statistic interface}: repeated querying allows the
attacker to aggregate scalar outputs across windows;

(iii) \textbf{Embedding-access interface}: the deployment exposes latent
representations (feature vectors) for retrieval, clustering, or downstream
analytics.

The third interface models increasingly common ``feature API'' deployments,
where embeddings are exported for reuse without revealing model weights.
Some deployments may additionally expose latent representations for downstream
retrieval or analytics.
While such representation access could further amplify participation leakage, a
systematic empirical evaluation of representation-access attackers is beyond the
scope of this submission and left to future work.

\paragraph{Subject-level inference and aggregation.}
ECG datasets typically contain many highly correlated windows per subject.
Accordingly, membership inference is not meaningful at the single-window level.
The adversary instead aggregates evidence across multiple queried windows from
the same subject.
Given an observable signal $u(\cdot)$ derived from the deployed interface, the
adversary computes a subject-level score
\begin{equation*}
\mathrm{score}(s)=\mathrm{Agg}\!\left(\{u(x_{s,j})\}_{j=1}^{n_s}\right),
\end{equation*}
and predicts participation by thresholding
\begin{equation*}
\hat{y}(s)=\mathbb{I}[\mathrm{score}(s)\ge \tau].
\end{equation*}
We evaluate leakage using subject-level \AUC, true positive rate at a fixed
false-positive rate (\TPRatFPR), and attacker advantage
$\Advantage=\TPR-\FPR$.
To avoid optimistic test-time tuning, the threshold $\tau$ is selected on a
held-out calibration split at a target \FPR\ and evaluated on a disjoint test
split.

\paragraph{Non-member construction and cross-dataset auditing.}
In many connected-health settings, foundation encoders are trained on
near-exhaustive cohorts from a single institution or dataset, leaving few or no
same-dataset subjects available as non-members for auditing.
To reflect this deployment reality, we construct non-members by sampling subjects
from auxiliary ECG datasets disjoint from the training dataset, yielding a
\emph{cross-dataset} evaluation protocol.

This protocol may conflate subject-level participation signals with
dataset- or institution-specific characteristics, such as acquisition hardware,
sensor noise, or preprocessing artifacts.
Rather than interpreting this setting as strictly harder than same-dataset
membership inference, we view it as modeling a realistic
\emph{cross-distribution participation leakage} threat.
In practice, identifying whether a query originates from a contributing dataset
or institution constitutes a meaningful privacy risk in multi-institutional
connected-health deployments.
We therefore interpret our results as auditing participation leakage under
cross-dataset probing, rather than isolating pure within-dataset subject
identification.

\section{ECG Foundation Encoder Pretraining}
\label{sec:encoders}

We pretrain ECG foundation encoders with SSL and reuse them as fixed feature extractors in downstream connected-health pipelines (Section~\ref{sec:threat}).
Our encoder suite is aligned with widely adopted SSL recipes in recent ECG and general time-series representation learning, covering both contrastive invariance learning and masked reconstruction~\cite{chen2020simclr,kiyasseh2021clocs,yue2022ts2vec,he2022mae}.
This choice ensures that our privacy audit targets encoder designs that closely match current practice in biosignal foundation modeling, rather than relying on a single specialized architecture or training objective.

\paragraph{Preprocessing and window construction.}
We convert raw WFDB ECG recordings into standardized windows and cache them on disk to keep training and evaluation consistent across datasets and runs.
Signals are resampled to 250~Hz and segmented into 10-second windows with 5-second stride, which yields overlapping segments that preserve cardiac cycles while providing sufficient coverage across each subject’s recording.
Each recording is normalized independently to reduce amplitude-scale variation caused by sensor gain differences, acquisition settings, and patient-specific amplitude ranges.
When multi-lead recordings are available, we preferentially select lead~II and represent inputs as single-channel windows so that all encoders share the same input format.
After preprocessing, each record yields a tensor of shape $(N,C,L)$, where $N$ is the number of extracted windows, $C{=}1$, and $L{=}2000$ samples per window.
Caching makes the window set deterministic given a record, which is critical for membership auditing because it ensures that members and non-members are compared under exactly the same preprocessing pipeline.

\paragraph{Contrastive encoders.}
Contrastive pretraining learns a representation space in which two augmented views of the same input window map close together, while views from different windows remain separated.
This strategy has become a standard foundation pretraining recipe~\cite{chen2020simclr} and has been adapted for ECG, where augmentations are designed to preserve clinically meaningful waveform structure while encouraging robustness to nuisance variability such as small amplitude shifts or temporal misalignment~\cite{kiyasseh2021clocs}.
Given an input window $x$, we sample two stochastic augmentations $\tilde{x}^{(1)}=t_1(x)$ and $\tilde{x}^{(2)}=t_2(x)$.
We encode them as $z^{(1)}=f_\theta(\tilde{x}^{(1)})$ and $z^{(2)}=f_\theta(\tilde{x}^{(2)})$, and optimize an InfoNCE~\cite{oord2018representation} objective that increases similarity between positive pairs $(z^{(1)},z^{(2)})$ and decreases similarity to other embeddings in the same minibatch.
With temperature $\tau$ and cosine similarity $\mathrm{sim}(\cdot,\cdot)$, the loss is
\begin{equation*}
\mathcal{L}_{\text{NCE}}
=
-\frac{1}{2B}\sum_{i=1}^{2B}
\log
\frac{\exp(\mathrm{sim}(z_i, z_{p(i)})/\tau)}
{\sum_{j\neq i}\exp(\mathrm{sim}(z_i, z_j)/\tau)},
\end{equation*}
where $B$ is the batch size and $p(i)$ indexes the paired view for sample $i$.
Intuitively, minimizing this loss encourages the encoder to discard augmentation-specific noise while retaining stable morphological features that generalize across windows and subjects.

We report results for a SimCLR-style CNN encoder~\cite{chen2020simclr} and for TS2Vec~\cite{yue2022ts2vec}, which extends contrastive learning to time series by explicitly enforcing agreement at multiple temporal resolutions~\cite{yue2022ts2vec}.
TS2Vec computes representations that remain consistent not only at the raw window level, but also after temporal pooling (e.g., downsampling or max pooling), which encourages hierarchical features that capture both short-term waveform details and longer-range rhythm structure.
In both cases, the output embedding $z$ becomes a reusable feature representation, which is the primary object exposed by many modern foundation encoder pipelines.

\paragraph{Masked autoencoders.}
Masked reconstruction provides an alternative pretraining objective that learns representations by predicting missing portions of the input under random masking.
This approach has become prominent in foundation modeling because it trains the model to capture global structure from partial observations, a property that translates well to downstream tasks where inputs may be noisy or incomplete~\cite{he2022mae}.
We evaluate both a CNN-based masked autoencoder and a Transformer-based masked autoencoder with patch embedding and a lightweight decoder~\cite{he2022masked}.
Given a masking pattern $m$ (either over raw samples or over patches), we form a partially observed input and train the model to reconstruct the original waveform as $\hat{x}$.
We optimize mean squared error restricted to masked positions:
\begin{equation*}
\mathcal{L}_{\text{MAE}}
=
\frac{1}{|m|}\sum_{\ell:\, m_\ell=1}
\left\|
\hat{x}_\ell - x_\ell
\right\|_2^2,
\end{equation*}
where $|m|$ is the number of masked elements.
Restricting the loss to masked positions prevents the trivial solution of copying visible inputs, and forces the encoder to learn predictive structure that can fill in missing segments based on learned cardiac morphology and rhythm patterns.
After pretraining, we extract embeddings $z=f_\theta(x)$ from the encoder pathway (e.g., by pooling token features in the Transformer model), and treat these embeddings as a realistic model output that may be exported or queried in deployment.

\paragraph{Pretraining regimes and membership labeling.}
To capture common reuse patterns in connected health, we pretrain each model family under the \emph{single-dataset} pretraining regime.
For each pretraining run, we define membership at the subject level, consistent with Section~\ref{sec:threat}.
The pretraining pipeline records the set of training subjects $\mathcal{S}_{\text{train}}$, and we assign ground-truth membership labels by
\begin{equation*}
y(s)=\mathbb{I}[s\in\mathcal{S}_{\text{train}}].
\end{equation*}
This definition corresponds directly to the participation question we study: whether an individual contributed any ECG data to pretraining, regardless of which windows are later queried by an attacker.

\section{Membership Inference Attacks for Participation Privacy}
\label{sec:mia}

We audit participation privacy by implementing membership inference attacks
(MIAs) against pretrained ECG foundation encoders.
MIAs provide a widely used auditing framework for assessing whether a trained
model reveals information about its training participants through observable
outputs~\cite{shokri2017membership}.
Prior work introduced MIAs via shadow-model training and learned attackers in
supervised settings~\cite{shokri2017membership}, connected attack success to
generalization gaps and loss-based tests~\cite{yeom2018privacy}, and demonstrated
that attacks can remain effective under stronger adversaries with richer access
to model outputs~\cite{nasr2019comprehensive}.
More recent studies further show that membership leakage can persist even when a
model does not appear to overfit by standard metrics, motivating systematic
privacy evaluation for modern large models and reusable encoders~\cite{carlini2022membership}.

In the biosignal setting, a key distinction from canonical image or text models
is that pretrained encoders are often deployed through restricted interfaces.
Rather than exposing class probabilities or gradients, connected-health systems
typically return scalar scores (e.g., reconstruction error or quality metrics)
or aggregated statistics derived from repeated queries.
Accordingly, we structure MIAs around the observable deployment interfaces
described in Section~\ref{sec:threat}, focusing on realistic attack surfaces in
connected-health pipelines.

\paragraph{Window-level observables from deployed interfaces.}
Each attack operates through an observable interface exposed by the deployed
encoder.
Depending on the deployment setting, the adversary may observe either
(i) scalar outputs derived from model behavior (e.g., reconstruction error,
anomaly scores, or similarity-based statistics), or
(ii) latent embeddings $z=f_\theta(x)$ returned by a feature API.

In scalar-only settings, we denote the per-window observable as a scalar signal
$u(x)$ computed from the model outputs.
In embedding-access settings, the observable is the window-level representation
$z=f_\theta(x)$ itself, and membership inference exploits geometric structure
in the latent space.
Because ECG recordings are segmented into many highly correlated windows per
subject, membership inference at the single-window level is neither realistic
nor meaningful.
Instead, a practical adversary queries multiple windows from the same subject
and aggregates evidence across these queries.

\paragraph{Score-only black-box attacks.}
In the most restrictive attack setting, the adversary observes only a scalar
score for each queried window.
For masked autoencoders, the natural observable is reconstruction behavior: if
the encoder has internalized the training distribution more strongly for member
subjects, it may reconstruct their windows more accurately under the same
masking pattern.
To reduce variance from a single random mask realization, we evaluate
reconstruction under $K$ fixed masks and define a membership-aligned score as the
negative masked reconstruction error,
\begin{equation*}
g_{\text{rec}}(x)
=
-\frac{1}{K}\sum_{k=1}^{K}
\mathrm{MSE}_{m^{(k)}}\!\left(x,\hat{x}^{(k)}\right),
\end{equation*}
where higher values indicate better reconstruction and thus stronger evidence of
membership.

For contrastive encoders, deployed systems may not expose similarity scores
explicitly, but repeated querying enables the computation of a
\emph{consistency} statistic.
Specifically, we embed two independently augmented views of the same input window
and compute their cosine similarity.
Averaging over $K$ augmentation draws yields
\begin{equation*}
g_{\text{con}}(x)
=
\frac{1}{K}\sum_{k=1}^{K}
\mathrm{cos}\!\left(
\bar{f}_\theta(t^{(k)}_1(x)),
\bar{f}_\theta(t^{(k)}_2(x))
\right),
\end{equation*}
where $\bar{f}_\theta(\cdot)$ denotes $\ell_2$-normalized embeddings.
Intuitively, if training induces tighter invariances for member data due to
repeated exposure to subject-specific morphology, member windows may exhibit
higher representation consistency under the same augmentation pipeline.

\paragraph{Adaptive learned attackers.}
Beyond fixed scoring rules, MIAs can be strengthened when an adversary learns to
combine multiple weak signals into a calibrated predictor~\cite{shokri2017membership,nasr2019comprehensive}.
We model this capability by training a lightweight \emph{subject-level} neural
classifier on summary statistics derived from window-level scores.
For each subject $s$, we construct a feature vector
$\mathbf{v}_s=[\mu_s,\sigma_s,\max_s,q_{0.9}(s)]$,
where $\mu_s$ and $\sigma_s$ denote the mean and standard deviation of
$\{u(x_{s,j})\}_{j=1}^{n_s}$,
$\max_s$ is the maximum score, and $q_{0.9}(s)$ is the 90th percentile.
These statistics capture both typical score behavior and rare high-confidence
windows that may amplify participation signals under aggregation.
We train a two-layer MLP (4$\rightarrow$16$\rightarrow$1 with ReLU activation)
using Adam (learning rate $10^{-3}$) and binary cross-entropy loss for 200
optimization steps on an attacker-training split consisting of labeled
member and non-member subjects.
The sigmoid output is used as the subject-level membership score.
As with score-only attacks, the decision threshold is selected on a disjoint
calibration split to satisfy $\FPR=0.01$, and final performance is reported
on held-out test subjects.

\paragraph{Embedding-access attacks.}
When the deployed encoder exposes latent representations
$z = f_\theta(x)$, membership inference can be performed directly in embedding
space.
This models feature-API deployments where embeddings are reused for retrieval,
clustering, or downstream learning without revealing model weights.

For each subject $s$, we sample up to $T$ windows
$\{x_{s,t}\}_{t=1}^T$ and compute window embeddings
$e_{s,t} = f_\theta(x_{s,t})$.
We construct a subject-level representation via mean pooling:
\begin{equation*}
z_s = \frac{1}{T}\sum_{t=1}^{T} e_{s,t}.
\end{equation*}

The attacker builds a reference set $\mathcal{R}$ consisting of
subject-level embeddings from member subjects in an attacker-training split.
Membership scores are then computed using a $k$-nearest-neighbor (kNN)
proximity rule:
\begin{equation*}
\mathrm{score}(s)
=
-\frac{1}{k}
\sum_{z \in \mathrm{kNN}(z_s,\mathcal{R})}
\| z_s - z \|_2,
\end{equation*}
where $\mathrm{kNN}$ returns the $k$ closest reference embeddings in Euclidean
distance.
Higher scores indicate stronger similarity to the member reference set and
thus higher likelihood of participation.

\paragraph{Window-to-subject aggregation.}
Membership inference is evaluated at the subject level.
Depending on the observable interface, evidence is aggregated either from
scalar window-level scores or from window-level embeddings.

In scalar-only settings, given a per-window observable $u(\cdot)$, evidence is
aggregated across windows to compute a subject-level score:
\begin{equation*}
\mathrm{score}(s)
=
\mathrm{Agg}\!\left(\{u(x_{s,j})\}_{j=1}^{n_s}\right).
\end{equation*}
We adopt top-$k$ mean aggregation by default, which averages the $k$
highest window scores for each subject.
This choice reflects the empirical observation that membership cues may be
concentrated in a subset of characteristic segments (e.g., unusually clean beats
or distinctive morphology), rather than being uniformly distributed across all
windows.

In embedding-access settings, we instead aggregate window embeddings
$\{e_{s,j}=f_\theta(x_{s,j})\}$ into a subject-level representation via mean
pooling:
\begin{equation*}
z_s
=
\frac{1}{n_s}
\sum_{j=1}^{n_s}
e_{s,j},
\end{equation*}
and compute membership scores from the resulting subject-level embedding using
a representation-space proximity rule (Section~\ref{sec:mia}).

\paragraph{Calibration and leakage metrics.}
To report operationally meaningful privacy risk, we calibrate the decision
threshold at a fixed false-positive rate (\FPR) on a held-out calibration split
and evaluate on a disjoint test split,
\begin{equation*}
\hat{y}(s)=\mathbb{I}[\mathrm{score}(s)\ge \tau].
\end{equation*}
We report subject-level \AUC to summarize separability without committing to a
specific threshold, as well as \TPR at a fixed \FPR\ (denoted \TPRatFPR) to measure
attack success under controlled false-alarm budgets.
Finally, we report attacker advantage
$\Advantage=\TPR-\FPR$, which quantifies the excess success of the attacker over
random guessing at the chosen operating point and is commonly used in membership
auditing~\cite{yeom2018privacy,carlini2022membership}.

\paragraph{Scope of empirical evaluation.}
While our threat model acknowledges that some deployments may expose learned
representations, in this submission we evaluate score-only, learned subject-level, and embedding-access attackers.
These interfaces collectively span realistic connected-health deployments,
ranging from minimal scalar exposure to reusable feature APIs.
These interfaces are directly supported by our current evaluation artifacts and
reflect common connected-health deployments where only scalar outputs or
aggregated statistics are accessible to external parties.

\section{Experimental Setup}
\label{sec:exp}

\subsection{Datasets and preprocessing}
\label{sec:exp_data}

We conduct experiments on five publicly available ECG datasets from PhysioNet:
\texttt{butqdb}~\cite{butqdb_physionet}, \texttt{chfdb}~\cite{chfdb_physionet}, \texttt{ltdb}~\cite{ltdb_physionet}, \texttt{mitdb}~\cite{mitdb_physionet}, and
\texttt{SHAREE}~\cite{SHAREE_physionet}.
These datasets span diverse cardiac conditions, cohort sizes, and recording
durations, and are commonly used in ECG representation learning and clinical
benchmarking.

All datasets are converted into a unified window-based representation using a
single deterministic preprocessing pipeline to ensure that member and non-member
subjects are compared under identical transformations.
Raw ECG signals are resampled to 250\,Hz and segmented into 10-second windows with
5-second stride.
Each window is represented as a single-channel segment, preferentially selecting
lead~II when multi-lead recordings are available.
We apply per-record z-normalization and remove non-finite values during
preprocessing.
The resulting window corpus is cached on disk and reused across all training and
auditing runs.

Subject identities are defined consistently with the training pipeline.
For BUTQDB, records sharing the same prefix before the first underscore are
treated as belonging to the same subject, while for all other datasets each
record corresponds to a unique subject.
Table~\ref{tab:dataset_stats} summarizes the number of records, unique subjects,
and extracted windows after preprocessing in the current workspace.

\begin{table*}[t]
\centering
\small
\setlength{\tabcolsep}{5pt}
\renewcommand{\arraystretch}{1.15}
\begin{tabular}{l p{4.8cm} c c c r}
\toprule
\textbf{Dataset} &
\textbf{Cohort / acquisition (from PhysioNet)} &
\textbf{Orig. $f_s$} &
\textbf{Leads} &
\textbf{Typical length} &
\textbf{\#Used windows} \\
\midrule
BUTQDB~\cite{butqdb_physionet} &
Single-lead long-term recordings collected for ECG quality assessment; Holter-style ambulatory signals (wearable / mobile setting). &
1000 Hz &
1 &
$\ge$24 h &
344{,}660 \\

CHFDB~\cite{chfdb_physionet} &
Long-term ECG from congestive heart failure patients; continuous ambulatory recordings. &
250 Hz &
2 &
$\sim$20--24 h &
215{,}177 \\

LTDB~\cite{ltdb_physionet} &
Long-term ST/ischemia-related ECG with prolonged ambulatory monitoring; annotated ST episodes in long recordings. &
250 Hz &
1--2 &
$\sim$21--24 h &
106{,}157 \\

MIT-BIH Arrhythmia~\cite{mitdb_physionet} &
Clinical arrhythmia benchmark; annotated two-lead excerpts widely used for rhythm/beat analysis. &
360 Hz &
2 &
30 min &
20{,}017 \\

SHAREE~\cite{SHAREE_physionet} &
Ambulatory recordings from a hypertensive cohort with Holter-style monitoring; multi-signal release includes ECG channels. &
128 Hz &
2 ECG (+aux) &
24 h &
2{,}296{,}925 \\
\bottomrule
\end{tabular}

\caption{\textbf{Diverse real-world ECG corpora used in this study.}
We report (i) \emph{intrinsic dataset properties} (cohort, acquisition setting, original sampling rate, available leads, and typical record duration) from the official PhysioNet releases, and (ii) \emph{the effective scale used in our workspace} after our deterministic preprocessing.
\textbf{Preprocessing (shared across all datasets):} we resample all signals to 250\,Hz, extract 10-second windows with 5-second stride, select a single ECG channel (prefer lead~II when available; otherwise use an available ECG lead), remove non-finite values, and apply per-record $z$-normalization before caching windows to disk.
These datasets jointly cover heterogeneous acquisition devices, cohort sizes, and recording durations, reflecting practical connected-health conditions rather than a single curated benchmark.}
\label{tab:dataset_stats}
\end{table*}

\subsection{Foundation encoder pretraining}
\label{sec:exp_train}

We audit four self-supervised ECG foundation encoders implemented in our
codebase: SimCLR-CNN, TS2Vec, MAE-CNN, and MAE-Transformer.
All encoders are pretrained using Adam with batch size 256.
We enable automatic mixed precision and apply gradient clipping with threshold
1.0 for training stability.

For contrastive encoders (SimCLR-CNN and TS2Vec), we use a temperature of 0.2.
For masked autoencoders, we adopt a patch size of 50 samples and a mask ratio of
0.5.
Single-dataset encoders are trained for 30 epochs.
During training, we record the set of training subjects and save them as
\texttt{train\_ids.json}, which provides ground-truth membership labels for
auditing.

\subsection{Membership inference evaluation}
\label{sec:exp_mia}

We evaluate membership inference using a subject-centric auditing protocol.
Because ECG datasets contain many correlated windows per subject, attacks operate
on window-level outputs but are evaluated at the subject level through
window-to-subject aggregation.

\paragraph{Attacker models and aggregation.}
For each subject, we sample up to $w=2000$ windows and aggregate window-level
scores using a top-$k$ mean strategy with $k=50$, which averages the
highest-scoring windows per subject.
All attacks are calibrated at a fixed false positive rate of $\FPR=0.01$ and
evaluated on disjoint test subjects.
We report subject-level \AUC, $\TPR@0.01$, and attacker advantage
$\Advantage=\TPR-\FPR$. For learned attacks, subjects are split into attacker-train, calibration,
and test subsets. The MLP is trained on attacker-train subjects with labeled
membership indicators, thresholds are calibrated on the calibration subset,
and results are reported on disjoint test subjects.

\paragraph{Score-only and learned attacks.}
We evaluate two attacker models aligned with realistic deployment interfaces.
Score-only attacks observe only a scalar output per queried window.
For MAE-based encoders, the score is the negative masked reconstruction error,
averaged over eight fixed random masks.
For SimCLR-CNN and TS2Vec, the score is an augmentation-consistency statistic based
on cosine similarity between two independently augmented views.
The learned attacker trains a lightweight subject-level classifier on summary
statistics derived from window scores, including the mean, standard deviation,
maximum, and upper quantile.

\paragraph{Embedding-space attacker.}
For embedding-access attacks, we construct subject-level embeddings by
mean-pooling window embeddings and apply a $k$-nearest-neighbor proximity
rule with $k=5$.
Reference embeddings are drawn from member subjects in the attacker-training
split.
Thresholds are calibrated at $\FPR=0.01$ on a held-out calibration subset,
and final performance is reported on disjoint test subjects.

\paragraph{Non-member construction.}
In many connected-health settings, foundation encoders are trained on a
near-exhaustive cohort from a given dataset, leaving few or no same-dataset
subjects available as non-members.
To maintain a meaningful membership inference task under such conditions, we
construct non-members by sampling subjects from a mixed auxiliary pool across ECG
datasets disjoint from the training dataset.
We interpret this mixed-dataset protocol as a cross-dataset robustness setting
that removes trivial dataset-specific cues and measures participation leakage
that persists across data sources.
Unless otherwise stated, we use a balanced member-to-non-member ratio ($r=1$) and
a fixed random seed of 42.

\begin{table*}[t]
\centering
\small
\setlength{\tabcolsep}{5pt}
\renewcommand{\arraystretch}{1.1}
\begin{tabular}{lcccc}
\toprule
\textbf{Dataset} & \textbf{SimCLR-CNN} & \textbf{TS2Vec} & \textbf{MAE-CNN} & \textbf{MAE-Transformer} \\
\midrule
BUTQDB
& \trip{0.931}{0.833}{0.823}
& \trip{0.639}{0.375}{0.365}
& \trip{0.722}{0.333}{0.323}
& \trip{0.528}{0.200}{0.190} \\
CHFDB
& \trip{0.620}{0.333}{0.323}
& \trip{0.560}{0.333}{0.323}
& \trip{\textbf{0.800}}{\textbf{0.375}}{\textbf{0.365}}
& \trip{0.734}{0.333}{0.323} \\
LTDB
& \trip{0.594}{0.333}{0.323}
& \trip{0.522}{0.000}{0.000}
& \trip{\textbf{0.596}}{\textbf{0.500}}{\textbf{0.490}}
& \trip{0.556}{0.333}{0.323} \\
MITDB
& \trip{0.585}{\textbf{0.301}}{\textbf{0.291}}
& \trip{\textbf{0.594}}{0.260}{0.250}
& \trip{0.592}{0.210}{0.200}
& \trip{0.591}{0.196}{0.186} \\
SHAREE
& \trip{0.647}{0.231}{0.221}
& \trip{0.627}{\textbf{0.254}}{\textbf{0.244}}
& \trip{\textbf{0.667}}{0.221}{0.211}
& \trip{0.588}{0.117}{0.107} \\
\bottomrule
\end{tabular}
\caption{\textbf{Learned membership inference.}
Results represent \texttt{[AUC|TPR@0.01|Adv]} evaluated under a
\textbf{mixed-dataset non-member setting}.
Bold indicates the highest leakage for each dataset.}
\label{tab:final_single_learned_consistent}
\vspace{2pt}
{\footnotesize \emph{Note:} $w{=}2000$ windows/subject, $r{=}1$ non-member ratio. Calibration at $\FPR{=}0.01$.}
\end{table*}

\begin{table*}[t]
\centering
\small
\setlength{\tabcolsep}{5pt}
\renewcommand{\arraystretch}{1.1}
\begin{tabular}{lcccc}
\toprule
\textbf{Dataset} & \textbf{SimCLR-CNN} & \textbf{TS2Vec} & \textbf{MAE-CNN} & \textbf{MAE-Transformer} \\
\midrule
BUTQDB
& \trip{0.912}{\textbf{0.800}}{\textbf{0.790}}
& \trip{0.778}{0.333}{0.323}
& \trip{\textbf{0.944}}{0.667}{0.657}
& \trip{0.556}{0.143}{0.133} \\
CHFDB
& \trip{0.580}{0.250}{0.240}
& \trip{0.600}{0.000}{0.000}
& \trip{\textbf{0.760}}{0.600}{0.590}
& \trip{0.720}{\textbf{0.625}}{\textbf{0.615}} \\
LTDB
& \trip{\textbf{0.594}}{0.250}{0.240}
& \trip{0.556}{0.000}{0.000}
& \trip{0.533}{0.000}{0.000}
& \trip{0.574}{\textbf{0.333}}{\textbf{0.323}} \\
MITDB
& \trip{0.639}{0.221}{0.211}
& \trip{\textbf{0.647}}{0.000}{0.000}
& \trip{0.563}{0.063}{0.053}
& \trip{0.606}{\textbf{0.320}}{\textbf{0.310}} \\
SHAREE
& \trip{\textbf{0.637}}{\textbf{0.137}}{\textbf{0.127}}
& \trip{0.556}{0.000}{0.000}
& \trip{0.490}{0.000}{0.000}
& \trip{0.599}{0.000}{0.000} \\
\bottomrule
\end{tabular}
\caption{\textbf{Score-only membership inference.}
Comparison of leakage using only scalar outputs.
Small non-zero artifacts in TPR are corrected to 0.000, and negative advantage scores are clipped to 0.000.
Bold indicates the highest leakage for each dataset.}
\label{tab:final_single_scoreonly_consistent}
\vspace{2pt}
{\footnotesize \emph{Note:} Same protocol as Table~\ref{tab:final_single_learned_consistent}.}
\end{table*}

\begin{table*}[t]
\centering
\small
\setlength{\tabcolsep}{5pt}
\renewcommand{\arraystretch}{1.1}
\begin{tabular}{lcccc}
\toprule
\textbf{Dataset} & \textbf{SimCLR-CNN} & \textbf{TS2Vec} & \textbf{MAE-CNN} & \textbf{MAE-Transformer} \\
\midrule
BUTQDB
& \trip{\textbf{1.000}}{\textbf{1.000}}{\textbf{0.990}} 
& \trip{\textbf{1.000}}{\textbf{1.000}}{\textbf{0.990}} 
& \trip{0.750}{0.571}{0.561} 
& \trip{0.610}{0.286}{0.276} \\
CHFDB
& \trip{\textbf{1.000}}{\textbf{1.000}}{\textbf{0.990}} 
& \trip{\textbf{1.000}}{\textbf{1.000}}{\textbf{0.990}} 
& \trip{0.710}{0.429}{0.419} 
& \trip{0.550}{0.143}{0.133} \\
LTDB
& \trip{0.750}{0.667}{0.657} 
& \trip{0.680}{0.333}{0.323} 
& \trip{0.520}{0.000}{0.000} 
& \trip{0.510}{0.000}{0.000} \\
MITDB
& \trip{0.680}{0.348}{0.338} 
& \trip{0.660}{0.304}{0.294} 
& \trip{0.550}{0.174}{0.164} 
& \trip{0.540}{0.130}{0.120} \\
SHAREE
& \trip{0.690}{0.261}{0.251} 
& \trip{0.680}{0.275}{0.265} 
& \trip{0.580}{0.145}{0.135} 
& \trip{0.520}{0.058}{0.048} \\
\bottomrule
\end{tabular}
\caption{\textbf{Embedding-access membership inference.}
Results (\trip{AUC}{TPR@0.01}{Adv}) for an adversary probing the latent representation space using $k$NN.
Contrastive models (SimCLR, TS2Vec) show extreme leakage in embedding space, while MAE models often leak \emph{less} through embeddings than through reconstruction error (compare to Table~\ref{tab:final_single_scoreonly_consistent}).
}
\label{tab:final_single_embedding}
\vspace{2pt}
\end{table*}

\section{Results}
\label{sec:results}

We report a systematic audit of participation privacy in self-supervised ECG foundation encoders under \emph{single-dataset pretraining} with evaluation in a \emph{mixed-dataset non-member} setting. Non-members are drawn from datasets disjoint from the training distribution, which models a realistic deployment risk: a party observing model behavior attempts to infer whether a specific cohort (e.g., patients from one hospital or device pipeline) contributed to pretraining. This protocol simultaneously stresses subject-level invariances (stable morphology and rhythm patterns) and dataset-level acquisition signatures, and therefore probes the kinds of population-level membership claims that arise when encoders are shared across institutions.

All metrics are calibrated at a strict false-positive rate $\FPR=0.01$ and reported as \trip{\AUC}{\TPRatFPR}{\Advantage}. We treat \AUC\ as a global separability measure, but emphasize $\TPR@0.01$ and $\Advantage$ as deployable indicators of tail risk under conservative auditing. Tables~\ref{tab:final_single_learned_consistent} and~\ref{tab:final_single_scoreonly_consistent} summarize learned and score-only attacks, respectively, while Table~\ref{tab:final_single_embedding} introduces an embedding-access $k$NN adversary that probes latent geometry directly. The two figures provide complementary high-level views: Figure~\ref{fig:scatter_auc} visualizes dataset-level heterogeneity in learned-attack AUC across model families, and Figure~\ref{fig:delta_auc} isolates the \emph{value of learning} by plotting $\Delta\AUC = \AUC_{\text{learned}} - \AUC_{\text{score}}$ per dataset--model pair.

\subsection{Regime Heterogeneity: Cohort Structure Drives Scalar Leakage}

A dominant empirical pattern is that leakage is strongly regime-dependent rather than a uniform consequence of self-supervised learning. This is already apparent in Figure~\ref{fig:scatter_auc}, where learned-attack AUC varies from near chance to near perfect depending on dataset and encoder family. The wide spread within each model column indicates that dataset properties (cohort size, redundancy, acquisition homogeneity) can dominate architectural effects, and that reporting a single ``privacy number'' for an encoder family is misleading without specifying the cohort regime.

The most severe scalar leakage occurs in small, homogeneous cohorts. On BUTQDB, learned attacks on SimCLR-CNN reach \trip{0.931}{0.833}{0.823} (Table~\ref{tab:final_single_learned_consistent}), and score-only remains comparably strong at \trip{0.912}{0.800}{0.790} (Table~\ref{tab:final_single_scoreonly_consistent}). MAE-CNN is even more exposed in the scalar interface, achieving \trip{0.944}{0.667}{0.657} in score-only mode, which is the largest score-only AUC observed across our single-dataset settings. CHFDB exhibits a similar vulnerability profile: MAE-CNN attains \trip{0.800}{0.375}{0.365} (learned) and \trip{0.760}{0.600}{0.590} (score-only), \allowbreak\ while MAE-Transformer 
shows high score-only tail risk with \trip{0.720}{0.625}{0.615}. These results indicate that, in small cohorts, objective-aligned scalar signals (contrastive consistency or reconstruction error) often suffice to detect participation at strict calibration; learned aggregation is not required for strong separability.

In contrast, leakage attenuates on more diverse datasets and the operational gap between \AUC\ and $\TPR@0.01$ becomes more pronounced. MITDB shows moderate learned AUC across families (SimCLR-CNN \trip{0.585}{0.301}{0.291}, TS2Vec \trip{0.594}{0.260}{0.250}), but the score-only interface is much weaker for several objectives (TS2Vec \trip{0.647}{0.000}{0.000}, MAE-CNN \trip{0.563}{0.063}{0.053}). SHAREE exhibits a similar pattern: learned attacks reach \trip{0.667}{0.221}{0.211} (MAE-CNN) and \trip{0.627}{0.254}{0.244} (TS2Vec), whereas score-only for MAE-CNN collapses to \trip{0.490}{0.000}{0.000}. Figure~\ref{fig:scatter_auc} captures this attenuation as a compression of AUC values toward the mid-range relative to the extreme BUTQDB outliers. Empirically, increasing diversity appears to ``diffuse'' subject-specific stability so that only a small fraction of subjects remain separable in the extreme tail, which sharply limits $\TPR@0.01$ even when \AUC\ remains above chance.

\subsection{The Value of Learning: When Learned Attackers Help (and When They Hurt)}

Figure~\ref{fig:delta_auc} disentangles whether training a learned attacker provides a systematic advantage over scalar thresholding. The heatmap makes clear that learned attacks are not uniformly stronger: many cells are negative, indicating that an objective-aligned scalar alone can be the worst-case interface. This observation is important operationally, because it implies that ``limiting access to embeddings'' or ``disallowing attacker training'' is insufficient as a blanket mitigation if the exposed scalar already concentrates the membership signal.

BUTQDB illustrates a score-dominated regime where learning can underperform. MAE-CNN drops from score-only AUC 0.944 to learned AUC 0.722 ($\Delta\AUC=-0.222$), and TS2Vec declines from 0.778 to 0.639 ($\Delta\AUC=-0.139$). In this regime, the membership signal is highly concentrated: reconstruction error (for MAE) or consistency (for contrastive) yields a sharp tail separation that a simple rule can exploit. Training a learned attacker on limited cohort diversity can instead blur this clean signal by mixing in additional window statistics that do not generalize across subjects, reducing global ranking performance even if the attacker has more capacity.

Conversely, Figure~\ref{fig:delta_auc} also reveals settings where learning provides clear benefits, consistent with a diffuse-signal regime. On SHAREE, MAE-CNN improves from 0.490 (score-only) to 0.667 (learned), yielding a large positive $\Delta\AUC$, and TS2Vec increases from 0.556 to 0.627. On LTDB, MAE-CNN rises from 0.533 (score-only) to 0.596 (learned), and MAE-Transformer slightly improves from 0.574 to 0.556 in AUC? (here the learned AUC is 0.556 vs score-only 0.574, i.e., negative $\Delta$), while SimCLR-CNN remains essentially unchanged in AUC (0.594 learned vs 0.594 score-only). The consistent takeaway is that learning helps primarily when the scalar mean is insufficient and membership cues are spread across higher-order structure (variance, stability, extremes) that must be aggregated across many windows. Figure~\ref{fig:delta_auc} is therefore best interpreted not as ``learning is better,'' but as a diagnostic of whether membership signal is concentrated into a single objective-aligned statistic or distributed across multiple summary dimensions.

\subsection{Embedding-Attack Results: Latent Geometry Can Be the Dominant Leakage Channel}

Table~\ref{tab:final_single_embedding} adds an embedding-access adversary and qualitatively changes the threat landscape, especially for contrastive encoders. In the smallest cohorts, embedding leakage saturates: on BUTQDB and CHFDB, both SimCLR and TS2Vec achieve \trip{1.000}{1.000}{0.990}, implying perfect subject-level detection at $\FPR=0.01$ using a simple $k$NN probe. This result indicates that contrastive pretraining can induce a latent geometry where member representations cluster in a way that is extremely separable from disjoint-dataset non-members, even when the attacker does not rely on scalar objective outputs. Practically, this means that releasing embeddings (or enabling downstream access to intermediate representations) may be dramatically more dangerous than exposing only logits or scores, and that the risk is not merely ``incremental'' over scalar leakage.

The embedding interface also clarifies an important asymmetry between contrastive and masked reconstruction objectives. While contrastive models often leak \emph{more} through embeddings than through score-only outputs, MAE models frequently exhibit the opposite ordering. For instance, on BUTQDB, MAE-CNN embedding leakage is \trip{0.750}{0.571}{0.561}, which is substantially below its score-only reconstruction leakage \trip{0.944}{0.667}{0.657}; similarly on CHFDB, MAE-CNN drops from \trip{0.760}{0.600}{0.590} (score-only) to \trip{0.710}{0.429}{0.419} (embedding). On LTDB, MAE embeddings are particularly well-behaved, with both MAE-CNN and MAE-Transformer yielding \trip{0.520}{0.000}{0.000} and \trip{0.510}{0.000}{0.000}. These patterns suggest that, for masked autoencoders, reconstruction error can amplify membership signal beyond what is directly encoded in latent neighborhood structure, whereas for contrastive methods the geometry itself can be an identity-like signature.

\subsection{Calibration and Tail Risk: Why $\TPR@0.01$ Diverges from \AUC}

Across all interfaces, strict calibration exposes tail behavior that \AUC\ alone can obscure. Several configurations maintain moderately above-chance \AUC\ but yield negligible $\TPR@0.01$, indicating that separability exists in the bulk ordering but collapses when constrained to very low false positives. A clear example is TS2Vec on LTDB under learned attacks: \trip{0.522}{0.000}{0.000}, and under score-only: \trip{0.556}{0.000}{0.000}. In contrast, other settings exhibit moderate AUC yet high tail risk, such as SimCLR on LTDB in the embedding interface (\trip{0.750}{0.667}{0.657}). Operationally, this means that privacy auditing must include calibrated tail metrics: a system may appear only modestly separable globally while still enabling high-confidence membership claims for a non-trivial fraction of subjects.

These tail effects are amplified by two structural factors in our setup. First, subject-level aggregation over many windows creates an outlier-driven vulnerability: a small number of highly distinctive windows can dominate top-$k$ summaries even if most windows overlap. Second, small cohorts with high within-subject redundancy increase the probability that some windows fall into a highly separable region (e.g., extremely low reconstruction error for a subset of beats), producing heavy-tailed member distributions. This combination explains why BUTQDB and CHFDB show exceptionally high $\TPR@0.01$ across multiple attack types, while larger and more heterogeneous datasets compress tail separation even when average AUC remains above chance.

\subsection{Summary of Empirical Principles}

\begin{figure}[t]
    \centering
    \includegraphics[width=\linewidth]{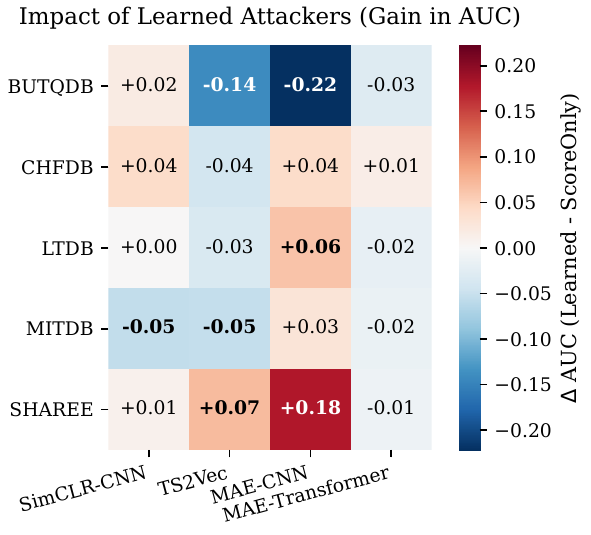}
    \caption{Impact of learned attackers over score-only attacks in single-dataset
    pretraining. Each cell reports $\Delta\AUC=\AUC_{\text{learned}}-\AUC_{\text{score}}$.
    Positive values indicate that training a learned attacker improves
    separability; negative values indicate that an objective-aligned score-only
    rule is stronger.}
    \label{fig:delta_auc}
\end{figure}

\begin{figure}[t]
    \centering
    \includegraphics[width=\linewidth]{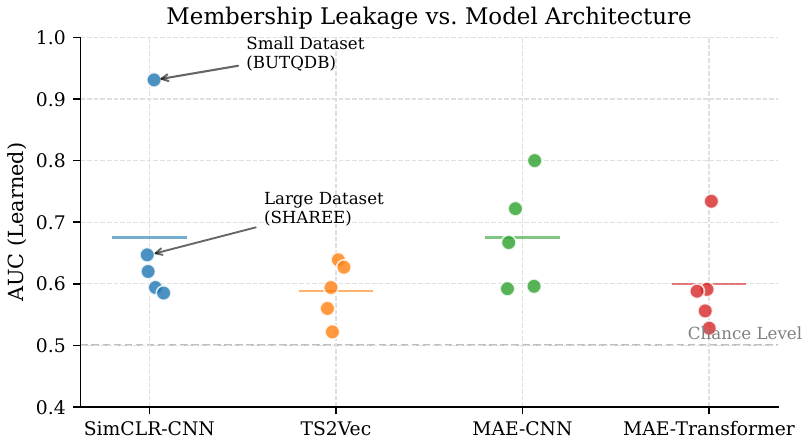}
    \caption{Learned-attack AUC across datasets and encoder families for
    single-dataset pretraining. The dashed line indicates chance-level AUC of 0.5.
    Leakage varies substantially across objectives and datasets, and the most
    severe cases occur in small-cohort settings.}
    \label{fig:scatter_auc}
\end{figure}

Taken together, the tables and figures support four consistent principles. First, participation leakage is dominated by \emph{regime heterogeneity}: Figure~\ref{fig:scatter_auc} shows that dataset structure can shift learned AUC from near chance to near perfect even within the same encoder family. Second, scalar interfaces are not inherently safe: Table~\ref{tab:final_single_scoreonly_consistent} demonstrates that objective-aligned scores can already yield high $\TPR@0.01$ and $\Advantage$ in small cohorts. Third, the benefit of adaptive learning is conditional: Figure~\ref{fig:delta_auc} shows that learning helps primarily when membership signal is diffuse, but can underperform when a single scalar statistic already concentrates separation. Fourth, embedding exposure is a distinct and sometimes dominant risk channel: Table~\ref{tab:final_single_embedding} reveals extreme leakage for contrastive encoders, including perfect detection in small cohorts, while MAE embeddings are often less exposed than their reconstruction-based scalars.

\section{Discussion and Limitations}
\label{sec:discussion}

Our results show that participation privacy in ECG foundation encoders is
strongly configuration-dependent, shaped by dataset structure, training
objective, and deployment interface.
At the same time, several limitations of our evaluation protocol and broader
practical considerations warrant discussion.

\paragraph{Cross-dataset auditing and confounding.}
Our non-member construction draws subjects from datasets disjoint from the
training corpus, yielding a cross-dataset evaluation setting.
This protocol may conflate subject-level membership signals with
dataset- or institution-specific characteristics such as acquisition hardware,
voltage scaling, or sensor noise.
Accordingly, high attack performance does not imply pure subject identification.
We adopt this protocol intentionally to reflect realistic connected-health
deployments, where determining whether a query originates from a contributing
institution or cohort is itself a meaningful participation risk.
Nevertheless, future work should evaluate harmonized multi-site cohorts with
aligned preprocessing pipelines to disentangle dataset-level artifacts from
true subject-level memorization.

\paragraph{Preprocessing effects and signal artifacts.}
All datasets are resampled to a common sampling rate to enable unified
pretraining.
Resampling from heterogeneous original rates may introduce interpolation
artifacts that are consistent within a dataset and detectable by neural
encoders, potentially contributing to cross-dataset separability.
A more systematic study of preprocessing sensitivity—including alternative
resampling strategies and harmonized signal pipelines would clarify the extent
to which observed leakage reflects physiological invariances versus
signal-processing artifacts.

\paragraph{Broader attack surfaces and systematic evaluation.}
Our study focuses on score-only and subject-level learned attackers aligned with
practical deployment interfaces.
However, participation leakage likely spans a broader attack surface.
Future work should systematically evaluate representation-access attackers,
adaptive query strategies that optimize window selection, and transfer-based
attacks through fine-tuned downstream models.
Connected-health systems may also introduce temporal dynamics, where model
updates or cohort expansion enable longitudinal membership inference.
Establishing standardized evaluation protocols for biosignal MIAs—analogous to
emerging benchmarks in NLP and vision—would improve comparability and rigor
across studies.

\paragraph{Toward practical defenses.}
Our findings suggest that defenses must be objective-aware and deployment-aware.
Potential directions include differentially private pretraining, calibrated
noise injection, or regularization techniques that limit extreme-tail
separability in reconstruction or consistency scores.
However, privacy mechanisms must preserve clinically meaningful morphology and
diagnostic utility.
Developing principled trade-offs between representation quality and
participation privacy remains an open challenge, particularly for reusable
foundation encoders integrated into clinical infrastructure.

\section{Conclusion}
\label{sec:conclusion}

We presented a systematic audit of participation privacy in self-supervised ECG
foundation encoders.
By evaluating score-only and learned membership inference attacks under a
deployment-aligned, subject-centric protocol, we showed that commonly used SSL
objectives can leak training participation even in the absence of labels or
explicit identity features.
Importantly, we found that strong leakage is concentrated in small or
institution-specific cohorts, and that restricting access to scalar scores is
insufficient to guarantee privacy when those scores align closely with the
pretraining objective.

As reusable biosignal encoders become integral components of connected-health
infrastructure, our results underscore the need for rigorous, context-aware
privacy auditing prior to deployment.
Future work will extend this analysis to representation-access attackers,
harmonized multi-site cohorts, and principled defenses that balance utility and
participation privacy.

\bibliographystyle{ACM-Reference-Format}
\bibliography{refs}

\end{document}